\documentclass[sigconf]{acmart}

\usepackage[table]{xcolor}
\AtBeginDocument{%
  }

\setcopyright{acmlicensed}
\copyrightyear{2018}
\acmYear{2018}
\acmDOI{XXXXXXX.XXXXXXX}
\acmConference[Conference acronym 'XX]{Make sure to enter the correct
  conference title from your rights confirmation email}{June 03--05,
  2018}{Woodstock, NY}
\acmISBN{978-1-4503-XXXX-X/2018/06}

\begin{document}

\title{GTSR: Subsurface Scattering Awared 3D Gaussians for Translucent Surface Reconstruction}

\author{Youwen Yuan}
\authornote{Both authors contributed equally to this research.}
\email{duskidew@qq.com}
\orcid{0009-0001-9063-882X}
\affiliation{%
  \institution{Xi'an Jiaotong University}
  \city{Xi'an}
  \state{Shaanxi}
  \country{China}
}

\author{Xi Zhao}
\affiliation{%
  \institution{Xi'an Jiaotong University}
  \city{Xi'an}
  \state{Shaanxi}
  \country{China}
}

\renewcommand{\shortauthors}{Yuan and Zhao}

\begin{abstract}
  Reconstructing translucent objects from multi-view images is a difficult problem. Previously, researchers have used differentiable path tracing and the neural implicit field, which require relatively large computational costs. Recently, many works have achieved good reconstruction results for opaque objects based on a 3DGS pipeline with much higher efficiency. However, such methods have difficulty dealing with translucent objects, because they do not consider the optical properties of translucent objects. In this paper, we propose a novel 3DGS-based pipeline (GTSR) to reconstruct the surface geometry of translucent objects. GTSR combines two sets of Gaussians, surface and interior Gaussians, which are used to model the surface and scattering color when lights pass translucent objects. To render the appearance of translucent objects, we introduce a method that uses the Fresnel term to blend two sets of Gaussians. Furthermore, to improve the reconstructed details of non-contour areas, we introduce the Disney BSDF model with deferred rendering to enhance constraints of the normal and depth. Experimental results demonstrate that our method outperforms baseline reconstruction methods on the NeuralTO Syn dataset while showing great real-time rendering performance. We also extend the dataset with new translucent objects of varying material properties and demonstrate our method can adapt to different translucent materials.
\end{abstract}


\begin{teaserfigure}
  \includegraphics[width=\textwidth]{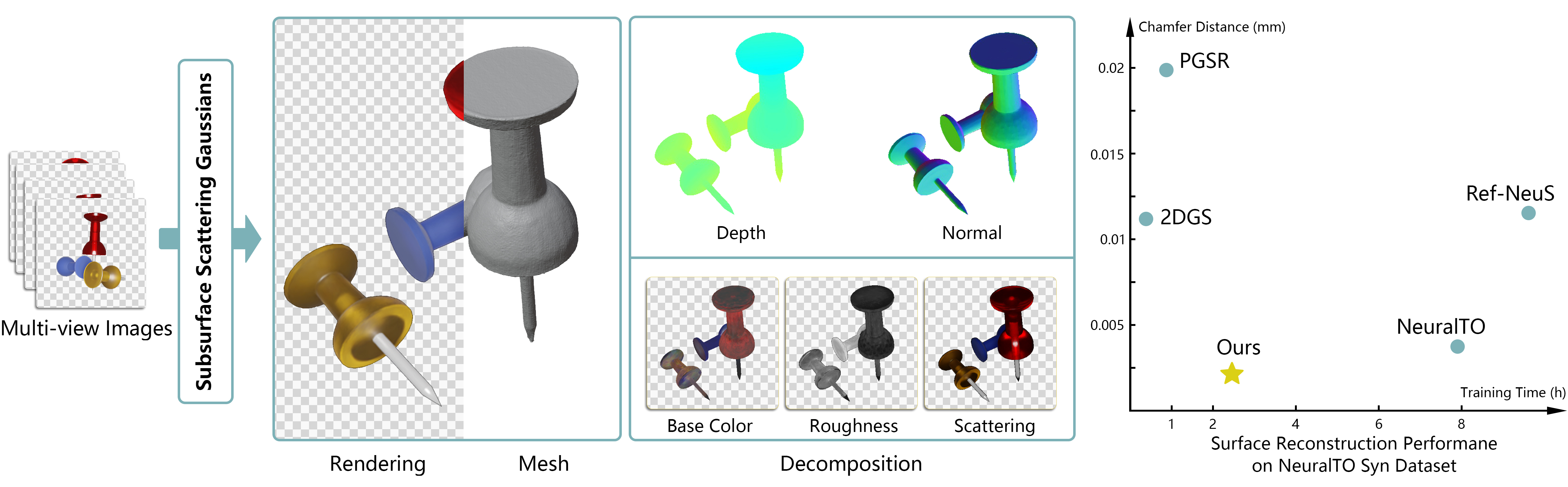}
  \caption{Our proposed pipeline takes the multi-view images of translucent objects as input, and aims to efficiently reconstruct the meshes by decomposing the geometry and material properties during optimization. Besides, our pipeline can also produce high-quality rendering results for novel views. Compared to NeRF-based and 3DGS-based surface reconstruction methods, our method can achieve the best surface reconstruction performance with efficient training process.}
  \label{fig:teaser}
\end{teaserfigure}


\maketitle

\section{Introduction}
Surface reconstruction and novel view synthesis (NVS) from multi-view images have been highly related and crucial tasks. NVS only focuses on the rendering quality of novel viewpoints. It may result in a good rendering effect but incorrect geometry. The surface reconstruction task is generally more challenging than NVS, especially for translucent objects. For opaque objects, only specular or diffuse reflection occurs when light hits their surfaces, which can be relatively easily simulated using classical lighting models. In contrast, light can penetrate the surface and enter the interior of translucent objects, leading to more complex optical phenomena such as refraction and scattering.

Many works try to solve the problem of reconstructing translucent objects. For example, a differentiable path tracing pipeline is used for inverse rendering of translucent objects \cite{recon_trans_obj_diff_rendering}. Concerning the heavy time consumption of differentiable path tracing, some works use neural rendering to tackle this problem \cite{li2023inverse, zheng2021neural}. And NeuralTO \cite{cai2024neuralto} proposes a surface reconstruction pipeline for translucent objects based on NeuS \cite{neus}, while it still requires a significant amount of time and involves a two-stage process.

Recently, 3D Gaussians \cite{3dgs} have made significant progress in novel view synthesis. By explicitly representing scenes with Gaussian ellipsoids and leveraging Gaussian splatting, 3DGS enables much faster training speeds compared to NeRF \cite{nerf} while supporting real-time rendering. Numerous works have introduced 3D Gaussians into surface reconstruction. With some targeted design, they can reconstruct high-quality surfaces while optimizing rendering results. For instance, 2DGS \cite{2dgs} employs 2D Gaussian kernels attached to object surfaces to represent geometry, achieving more accurate geometric representations. PGSR \cite{chen2024pgsr} utilizes flattened 3D Gaussian kernels for geometric representation and proposes a novel unbiased depth computation method. However, these methods cannot adapt well to translucent objects with complex optical properties. The main reason is that these methods lack a suitable rendering model to guide the correct optimization results.

In this paper, we propose a novel 3DGS-based pipeline (GTSR) to precisely reconstruct the surface geometry of translucent objects with more efficient training process. To achieve this target, we need to solve two difficulties.
First, when reconstructing translucent objects, the optimization objective for achieving higher-quality rendering may conflict with the requirements of surface reconstruction. Existing 3DGS-based surface reconstruction methods \cite{2dgs, sugar} suggest that to improve surface reconstruction accuracy, gaussian kernels need to align with the surface. However, due to subsurface scattering in translucent objects, the optimization process tends to generate numerous Gaussian kernels inside the object to fit the scattering colors. Meanwhile, the opacity of Gaussian kernels near the surface decreases to ensure the visibility of these interior Gaussian kernels. As a result, the 3D Gaussian kernels no longer remain tightly concentrated near the surface, compromising reconstruction fidelity. Second, we cannot use the multi-view photometric consistency to refine the surface geometry. This is because that the assumption that the color of a point on the surface does not change greatly under different views is not suitable for translucent objects.

To deal with the first problem, we use two sets of Gaussian kernels, surface and interior Gaussians, to represent the surface color and scattering color, respectively. By using the two independent sets, we can use flat Gaussian kernels to represent the surface geometry, which helps to improve the precision of the reconstructed mesh. Besides, we separate the representation of appearance color of translucent objects by using two sets of Gaussians, which can avoid surface Gaussians influenced by the scattering and make surface Gaussians focus on representing surface geometry. Then, to make the interior Gaussians visible without the need of reducing the opacity of surface Gaussians, we add the Fresnel term into the rendering pipeline to simulate the attenuation when the lights get through the surface. 

To deal with the second problem, we introduce Physically Based Rendering (PBR) based on 3DGS pipeline to enhance the constraints of normal and depth. By supervising the rendering results of PBR module, we can decompose the geometry and material properties from the input images during the training process, which serves to further refine the details of non-contour areas. Here, we combine specular, dielectric diffuse and scattering terms to represent the final color according to the Disney BSDF model \cite{disney_bsdf}. Meanwhile, we leverage the deferred rendering to realize our PBR module, which has lower computational complexity and provides a more stable optimization process.

Finally, we evaluated our proposed method on NeuralTO's \cite{cai2024neuralto} translucent object reconstruction dataset to assess its performance in surface reconstruction and novel view synthesis. Experimental results confirm that our method achieves the best reconstruction quality on this dataset while delivering excellent real-time rendering performance. And our approach requires less time than the state-of-the-art NeuS-based method. Additionally, we expanded the dataset by incorporating more translucent objects with varying materials to test the algorithm's adaptability to different materials. The results demonstrate that our method exhibits superior performance and higher stability when handling objects with diverse material properties.

Our main contributions can be summarized as follows:
\begin{enumerate}
  \item We propose a 3DGS-based surface reconstruction pipeline for translucent objects.
  \item By employing two distinct sets of Gaussians, surface and interior Gaussians, we eliminate the optimizing conflicts during the surface reconstruction of translucent objects.
  \item We use PBR with deferred rendering to decompose the geometry and material properties, further enhancing geometric constraints on the surface.
  \item We achieve state-of-the-art performance of surface reconstruction of translucent objects on the NeuralTO Syn dataset\cite{cai2024neuralto}. Additionally, we extend the dataset with objects of varying material properties, validating our method's adaptability across different translucent materials.
\end{enumerate}

\section{Related Work}
\textbf{NeRF-based Surface Reconstruction.}
NeRF \cite{nerf} has achieved great success in novel view synthesis, but its density-based representation lacks explicit geometric constraints. To address this, several works introduce geometry-aware implicit representations. VolSDF \cite{volsdf} and NeuS \cite{neus} employ signed distance fields (SDFs) and volume rendering to recover accurate surfaces, while UNISURF \cite{unisurf} adopts a neural occupancy field. Neuralangelo \cite{neuralangelo} leverages multi-resolution hash grids to capture fine geometric details. HF-NeuS \cite{hf-neus} decomposes the SDF into base and displacement components to better model different frequency bands. Ref-NeuS \cite{ref-neus} and NeRO \cite{nero} further consider reflective surfaces via anomaly detection or PBR modeling. To improve efficiency, NeuS2 \cite{neus2} accelerates training using hash encodings. Despite their high-quality reconstructions, NeRF-based methods generally require several hours of optimization.

\textbf{3DGS-based Surface Reconstruction.}
3DGS \cite{3dgs} significantly improves training efficiency through Gaussian splatting and has recently been extended to surface reconstruction. SuGaR \cite{sugar} extracts meshes from flattened Gaussians, while 2DGS \cite{2dgs} and Gaussian Surfels \cite{gs-surfels} represent surfaces using oriented 2D primitives. GS2Mesh \cite{gs2mesh} reconstructs RGB-D geometry via stereo rendering, and PGSR \cite{chen2024pgsr} proposes an unbiased depth rendering scheme for smoother optimization. Other methods combine 3DGS with SDFs, such as 3DGSR \cite{3dgsr}, NeuSG \cite{neusg}, and GSDF \cite{gsdf}, or define surfaces via opacity fields \cite{gof}. GS-2DGS \cite{gs-2dgs} introduces PBR for reflective surfaces. However, these methods are not designed for translucent objects and cannot explicitly model subsurface scattering.

\textbf{Inverse Rendering based on NeRF and 3DGS.}
Inverse rendering aims to recover material properties from multi-view images, often using PBR to reduce ambiguities caused by reflection. Ref-NeRF \cite{ref-nerf}, IRON \cite{iron}, TensoIR \cite{tensoir}, and NVDiffrec \cite{nvdiffrec} recover surface materials and lighting under different assumptions. Several works incorporate advanced material models, such as microflake volumes \cite{nemf} or learned light transport \cite{relighting-nerf}. With the rise of 3DGS, inverse rendering methods such as GIR \cite{gir}, GS-IR \cite{gs-ir}, GaussianShader \cite{gaussianshader}, and RelightableGS \cite{relightablegs} extend PBR to Gaussian representations, often using deferred rendering \cite{reflective-gs, deferred-reflection, deferredgs}. OLAT \cite{olat} and TransparentGS \cite{TransparentGS} further consider scattering or refraction effects. Nevertheless, these methods mainly rely on BRDF-based models and are designed for relighting rather than mesh reconstruction, making them unsuitable for translucent objects.

\textbf{Reconstruction for Translucent Objects.}
Reconstructing translucent objects remains challenging due to complex light transport. Early works rely on time-of-flight sensors \cite{shim2015recovering, lee2015skewed}, which limits applicability. Other approaches use differentiable rendering \cite{recon_trans_obj_diff_rendering, thin_trans_obj} or explicit material estimation \cite{yang2016inverse, li2023inverse, zheng2021neural}, but often suffer from high computational cost. NeuralTO \cite{cai2024neuralto} achieves high-quality translucent surface reconstruction based on NeuS. However, existing methods have not explored leveraging the efficiency and rendering quality of 3DGS for translucent object reconstruction.

\section{Preliminaries}

\textbf{3D Gaussian Splatting.} 3DGS\cite{3dgs} uses a set of 3D Gaussian kernels to represent a scene. Each Gaussian kernel $G_i$ can be represented as $G_i(x) = e^{-\frac{1}{2}(x-\mu_i)^{\top}\sum_i^{-1}(x-\mu_i)}$ via 3D covariance matrix and mean, where $x$ is a 3D point, $\mu$ is the center of the Gaussian kernel, covariance matrix $\Sigma = \mathrm{RSS}^{\top}\mathrm{R}^{\top}$ records the rotation $\mathrm{R}$ and scaling $\mathrm{S}$. During the rendering, each Gaussian kernel will be projected to a 2D Gaussian, which is parameterized by 2D mean $\mu_i^\prime$ and covariance $\Sigma^\prime_i$:
\begin{equation}
    \mu_i^\prime = \mathrm{KW[\mu_i, 1]^\top}, \quad \Sigma_i^\prime = \mathrm{JW\Sigma_i W^\top J^\top}
\end{equation}
where $\mathrm{J}$ is the Jacobian matrix of projection matrix, $\mathrm{W}$ and $\mathrm{K}$ are the transform and intrinsic matrix, respectively. Then, the opacity of each Gaussian kernel at pixel $u$ is $\alpha_i = opacity_i * G_i(u|\mu_i^\prime, \Sigma_i^\prime)$. Finally, the color $C$ of the pixel $u$ in the view direction $\omega_o$ can be calculated by alpha blending.

\textbf{PGSR for Surface Reconstruction.} PGSR\cite{chen2024pgsr} proposes a novel unbiased depth and achieves good surface reconstruction results. It flattens the 3D Gaussian kernels and treats them as planar surfaces. Then it uses the direction of the shortest axis of the Gaussians as the normal $n_i$ for each Gaussian. The normal map $\mathbf{N}$ can be rendered by:
\begin{equation}
    \mathbf{N} = \sum_{i=1}^N {R_c^\top}n_i \alpha_i  \prod\limits_k^{i-1} (1-\alpha_k)
\end{equation}
where $R_c$ is the rotation matrix from view space to world space of the camera. And PGSR proposes an unbiased depth rendering method. The distance between camera and the center of Gaussian kernel is $d_i = (R_c^\top(\mu_i - T_c)) \cdot R_c^\top n_i$.
Then the planar distance map $\mathbf{D}_{plane}$ can be rendered by:
\begin{equation}
    \mathbf{D}_{plane}= \sum_{i=1}^N d_i \alpha_i  \prod\limits_k^{i-1}(1-\alpha_k)
\end{equation}
Finally, the depth map $\mathbf{D}$ can be calculated by normal map and planar distance map:
\begin{equation}
    \mathbf{D}(p)=\frac{\mathbf{D}_{plane}(p)}{\mathbf{N}(p)\cdot r}
\end{equation}
where $p$ is a pixel in screen space, and $r$ is the view direction in view space.

\textbf{Diseny BSDF.} Recently, some other works \cite{neural_sss, neu_pre_sss} try to use neural networks to represent translucent objects and realize efficient subsurface scattering rendering. However, they need well-prepared neural models and still need numerous sampling rays. The Disney BSDF model\cite{disney_bsdf}, based on empirical observations, decomposes the complex subsurface scattering phenomenon into three manageable components, Specular BSDF, Dielectric BRDF and Subsurface Diffusion. Specular BSDF $f_s$ contains BRDF and BTDF. It can be computed by the following equation:
\begin{equation}
    f_s = \frac{F_r(i,o) GD}{4(i \cdot n)(o \cdot n)} + \frac{\eta(1-F_r(i,o, \eta))GD}{(1+\eta)^2(i \cdot n)(o \cdot n)}
\end{equation}
where $i$ and $o$ are the incident and outgoing light directions, $n$ is the normal, $F_r$ denotes Fresnel reflection, $G$ is the geometric attenuation term, $D$ is the normal distribution function, and $\eta$ is the relative index of refraction. 
Meanwhile, the dielectric BRDF that concerns the scattering is modified as:
\begin{equation}
    f_d = \frac{baseColor}{\pi}(1-0.5F_L)(1-0.5F_V) + f_{retro-ref}
\end{equation}
where $baseColor$ is one of material properties, $F_L=(1-cos \theta_l)^5$, $F_V=(1-cos \theta_v)^5$, $f_{retro-ref}$ is the retro-reflection component.

Subsurface diffusion is too complex and difficult to implement based on 3DGS rendering pipeline, so we only use a simplified approach. 

\begin{figure*}
    \includegraphics[width=\textwidth]{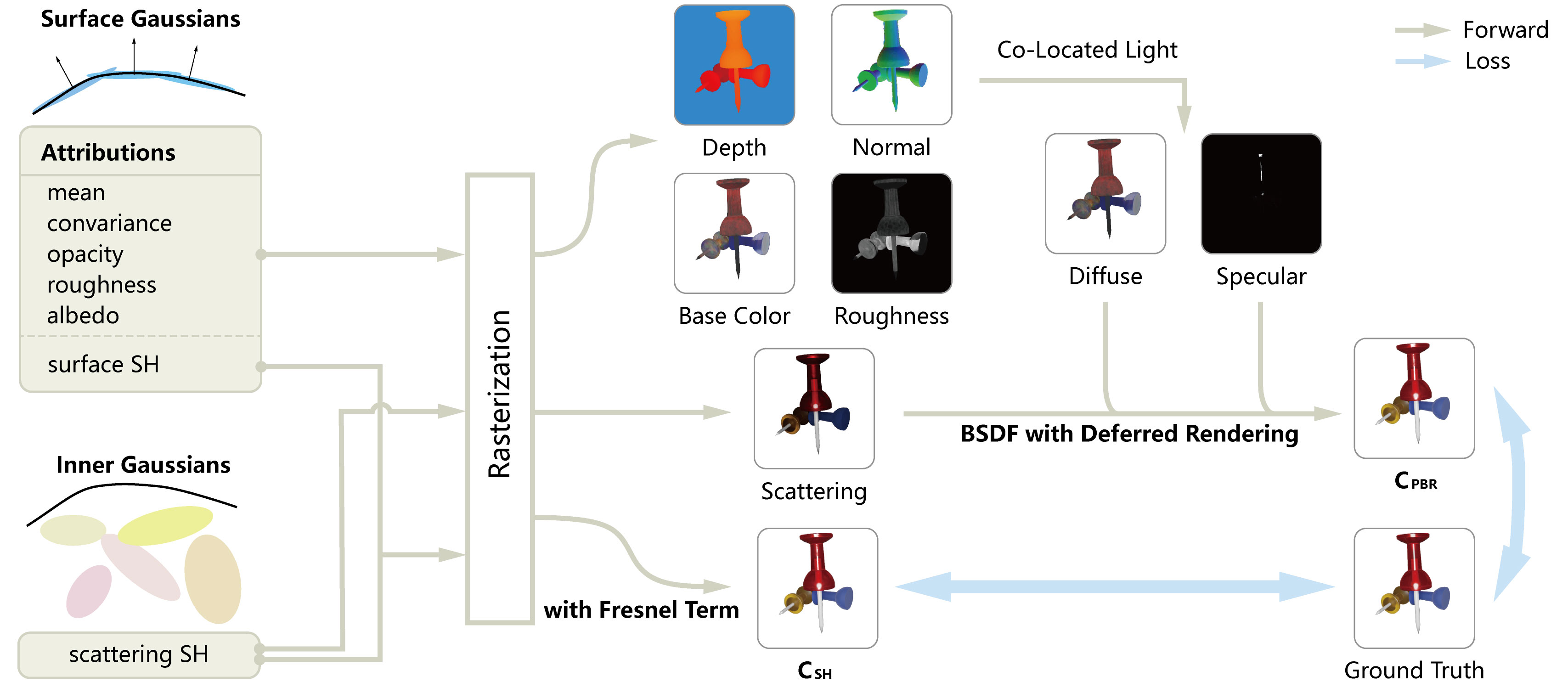}
    \caption{The overview of the surface reconstruction pipeline. We combine the surface and interior Gaussians by introducing a Fresnel term and get the first rendering result $C_{\mathrm{SH}}$. Then, we use surface Gaussians to brake several texture maps (depth, normal, base color and roughness), and leverage interior Gaussians to render the scattering map. After that, we obtain the second rendering result $C_{\mathrm{PBR}}$ by blending these maps based on Disney BSDF model in screen space. Finally, we use ground truth to supervise the two rendering results simultaneously.}
    \label{fig-overview}
\end{figure*}

\section{Method}
Given a group of multi-view images of translucent objects, we aim to achieve precise surface reconstruction with higher efficiency, and keep high-quality rendering performance. The overview of our novel pipeline is illustrated in Fig. \ref{fig-overview}. Firstly, we use two sets of Gaussian kernels to represent the appearance of translucent objects, surface Gaussians and interior Gaussians. The surface geometry is only represented by surface Gaussians. Then, we obtain the final rendering results by introducing a Fresnel term to merge surface and interior Gaussians. To further improve the details of reconstructed surfaces, we use PBR for translucent objects based on the 3DGS rendering pipeline and Disney BSDF model. Using deferred rendering, our PBR module can be easier to implement and requires less calculation overhead.  

\subsection{Surface and Interior Gaussians}
\begin{figure}
    \includegraphics[width=\linewidth]{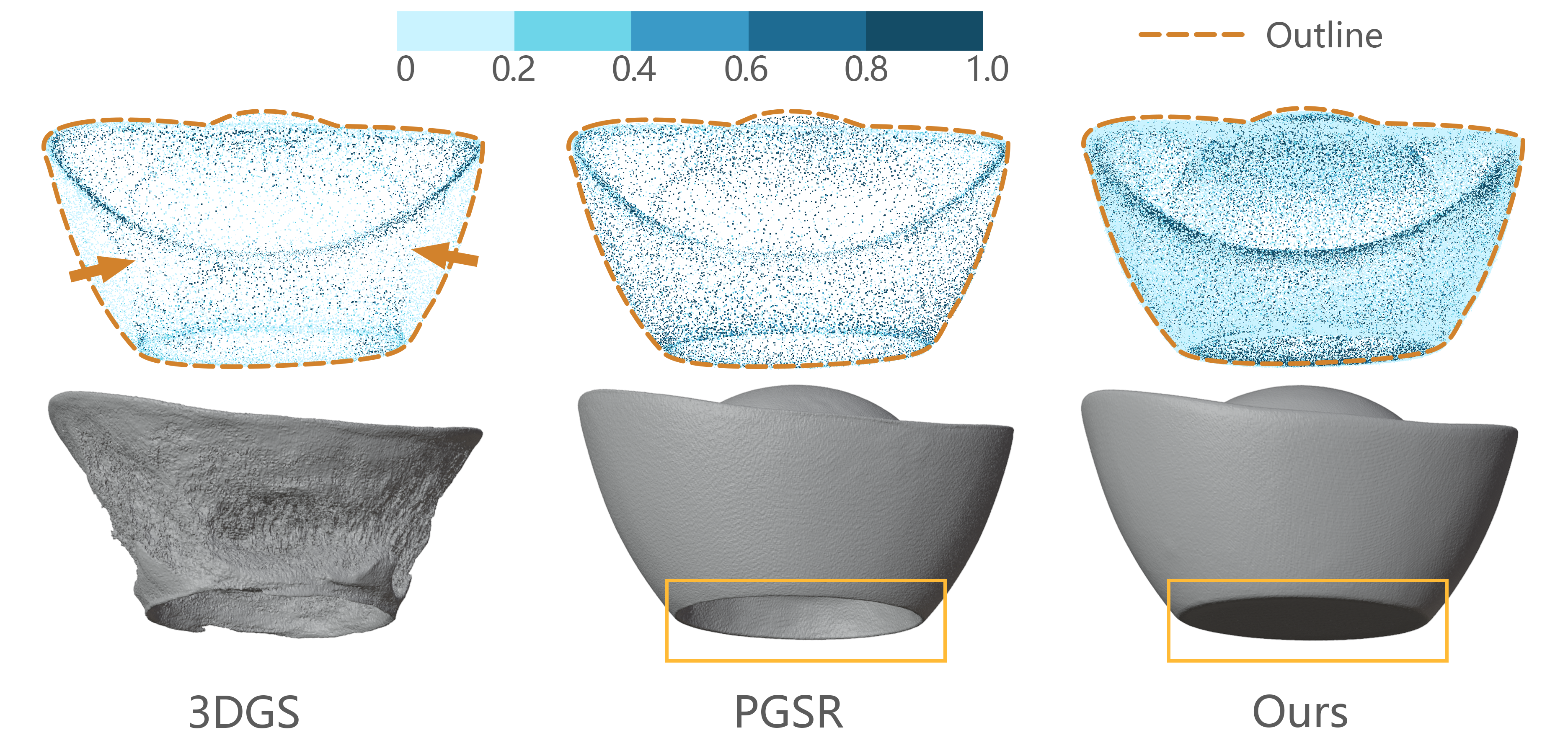}
    \caption{The visualization of optimization results for different methods. 
    The first row shows the opacity of Gaussian kernels. The second row shows the reconstructed meshes for different methods. The color banding at the top shows the opacity ranging from 0 to 1. Compared to other methods, our method can produce much denser and opaque surface Gaussians. The area of 3DGS's result pointed out by the arrows shows that many Gaussian kernels with high opacity are located inside the object.}
    \label{fig:surf-inner-gs}
\end{figure}

Previous 3DGS-based methods use a single set of Gaussian kernels to represent object appearance. When applied to translucent objects, this design tends to place many Gaussians inside the object to fit scattering effects, while surface Gaussians often have low opacity to allow interior visibility. Although such representations can produce acceptable novel view synthesis results, they are suboptimal for surface reconstruction. Accurate surface reconstruction requires dense, high-opacity Gaussians on the surface to generate reliable depth maps, which are crucial for TSDF-based mesh extraction \cite{newcombe2011kinectfusion}.

To simultaneously achieve high-quality rendering and accurate surface geometry, we introduce two sets of Gaussian kernels with different roles: (1) surface Gaussians, which are dense and have high opacity to represent surface geometry and color; and (2) interior Gaussians, which are located inside the object to model scattering effects. As shown in Fig. \ref{fig:surf-inner-gs}, our method produces denser, high-opacity Gaussians on the object surface compared to 3DGS and PGSR. In contrast, 3DGS assigns low opacity (often below 0.2) to most surface Gaussians while placing high-opacity Gaussians inside the object. As a result, our reconstructed meshes are significantly smoother, whereas baseline methods suffer from cavities caused by sparse, low-opacity surface Gaussians.

\subsection{Rendering with Fresnel Term}
\begin{figure}
    \centering
    \includegraphics[width=\linewidth]{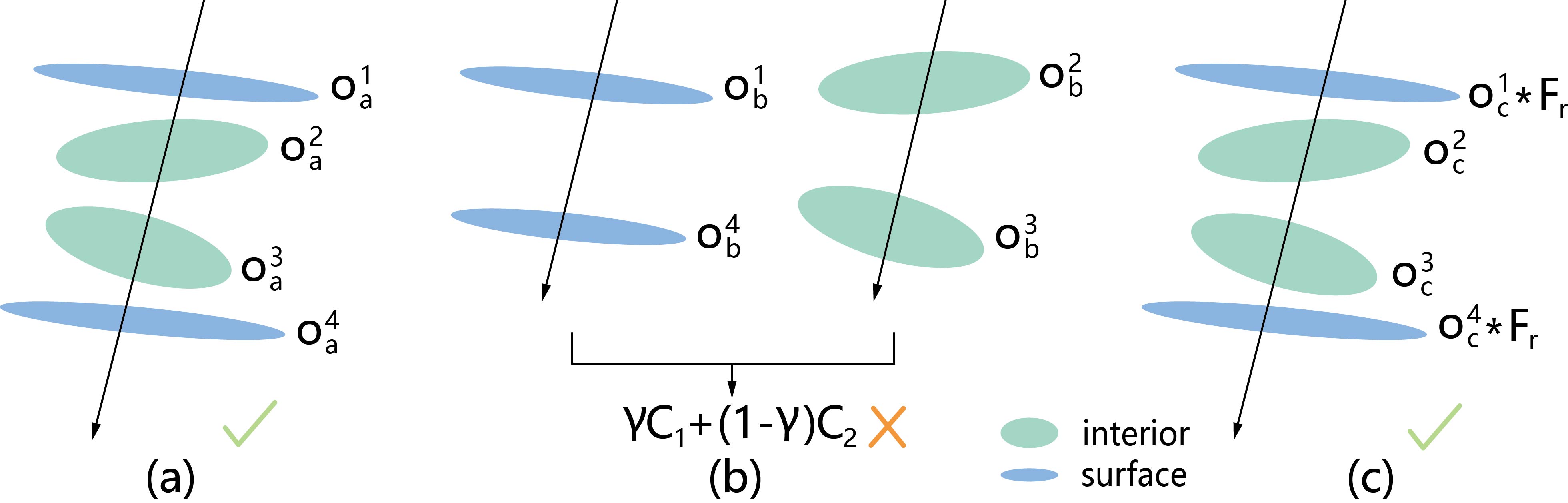}
    \caption{Different ways to blend the Gaussian kernels the light pass through. The notes beside the ellipses represent the opacity. (a): The method used in 3DGS. (b): Using surface and interior Gaussians to render $C_1$ and $C_2$, then blending $C_1$ and $C_2$ with a weight $\gamma$. NeuralTO adopts this method. (c): Embedding weights $F_r$ into 3DGS pipeline.}
    \label{fig:fresnel-term}
\end{figure}
After we use two sets of Gaussians to represent the appearance of translucent objects, the next key problem is how to combine the two sets of Gaussians during the rendering.

If we concatenate the two sets of Gaussians directly and send them into this rendering pipeline, the surface Gaussians still need to keep a low opacity to make the interior Gaussians visible after optimization. This can result in a bad surface geometry (Fig.\ref{fig:surf-inner-gs}). 
Another way is to use two sets of Gaussian kernels for rendering independently and merge their rendering results by weights, as illustrated in Fig.\ref{fig:fresnel-term} (b). However, in this method, the interior Gaussians will be sorted after all surface Gaussians, which is not the correct order for the view rays to pass through the object. 
So, in our method, we still combine the two sets of Gaussians and send them into the 3DGS pipeline to get the correct order. The difference is that we consider the attenuation phenomenon by multiplying a weight to the opacity whenever the view ray passes a surface Gaussian kernel (shown in Fig.\ref{fig:fresnel-term} (c)).

More specifically, we choose the Fresnel term as the weight. By multiplying by Fresnel term, a surface Gaussian will become more transparent when the view ray is nearly parallel to its orientation, which effectively simulates the characteristic that a translucent object's surface appears more transparent when viewed nearly perpendicularly to the line of sight. This view-dependent opacity can be formulated as:
\begin{equation}
    \alpha_i^\prime = \alpha_i * F_{r,i}
\end{equation}
where, the Fresnel term $F_{r,i}$ for each surface Gaussian $G_i$ can be computed by using Schlick's approximation\cite{schlick}:
\begin{equation}
    F_{r, i}=F_0 + (1-F_0)(1-n_i \cdot \omega_o)^5
\end{equation}
where, $F_0$ is the basic reflectance, we usually set it to 0.04. $n_i$ is the normal of surface Gaussian $G_i$, which is estimated by the direction of the surface Gaussians' shortest axis (similar to PGSR). $\omega_o$ is the view direction.

For all surface Gaussians, we use the spherical harmonic $\mathrm{SH}_{\mathrm{surf}}(\omega_o)$ to represent the surface color under view $\omega_o$. For the interior part, we also assign a spherical harmonic $\mathrm{SH}_{\mathrm{scatter}}$ to each interior Gaussian to represent the scattering. Finally, we can combine the surface and interior Gaussians to get the rendering result $C_{\mathrm{SH}}$ for the view ray $\omega_o$ by the method illustrated in Fig.\ref{fig:fresnel-term} (c). This can be formalized as:
\begin{equation}
    C_{\mathrm{SH}} =\sum_i^N c_i \alpha_i^\prime  \prod\limits_k^{i-1}(1-\alpha_k^\prime)
\end{equation}
where, $c_i$ is $\mathrm{SH}_{\mathrm{surf}}^i(\omega_o)$ for surface Gaussians and $\mathrm{SH}_{\mathrm{scatter}}^i(\omega_o)$ for interior Gaussians.
And $\alpha_i^\prime$ is $\alpha_i F_{r, i}$ for surface Gaussians and $\alpha_i$ for interior Gaussians.
$N$ is the total number of the surface Gaussians $G_{\mathrm{surf}}$ and interior Gaussians $G_{\mathrm{in}}$ that the view ray passes through. 

\subsection{BSDF with Deferred Rendering}
\begin{figure}
    \includegraphics[width=\linewidth]{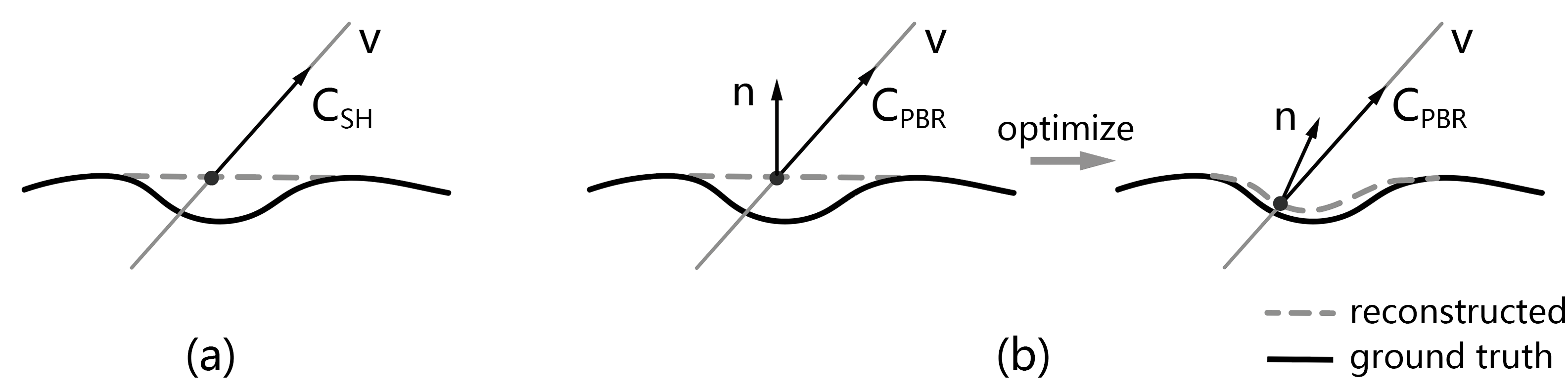}
    \caption{The different effects of optimizing $C_{\mathrm{SH}}$ and $C_{\mathrm{PBR}}$ on surface geometry. Different type of lines indicate the surface. (a): $C_{\mathrm{SH}}$ renders the appearance of the object via view-dependent color representation, spherical harmonics. The color of spherical harmonics is only related to the view direction, so the optimization results of rendering may be great while the surface geometry could be incorrect. (b): $C_{\mathrm{PBR}}$ needs correct normal to get the correct rendering results. Therefore, by supervising the rendering results, the normal can be optimized in the correct direction.}
    \label{fig:pbr-surf-geo}
\end{figure}


Most surface reconstruction methods rely on multi-view color consistency to improve geometry quality. For example, PGSR \cite{chen2024pgsr} introduces a multi-view photometric consistency regularization based on the assumption that a surface point has the same color across different view directions. While this assumption holds for opaque objects in datasets such as DTU \cite{dtu} and Tanks and Temples \cite{knapitsch2017tanks}, it breaks down for translucent objects, whose surface appearance can vary significantly across views due to specular reflection and subsurface scattering.

To address this limitation, we introduce an optimizable PBR module that explicitly incorporates depth and surface normals into rendering, enabling tighter geometric constraints during optimization. Our PBR module is based on the Disney BSDF model \cite{disney_bsdf}, which is widely used for rendering translucent materials. It consists of three components: Specular BSDF, Dielectric BRDF, and Subsurface Diffusion. In our formulation, the Specular BSDF and Dielectric terms are modeled using the geometry and material properties of surface Gaussians, while the Subsurface Diffusion term is represented by interior Gaussians. These components are efficiently combined using deferred rendering to produce the final result.
As illustrated in Fig. \ref{fig:pbr-surf-geo}, supervising only view-dependent rendering results $C_{\mathrm{SH}}$ may lead to incorrect geometry (Fig. \ref{fig:pbr-surf-geo}(a)). By contrast, our PBR module tightly couples rendering with surface geometry, enabling the optimized shape to better match the ground truth (Fig. \ref{fig:pbr-surf-geo}(b)), resulting in more accurate surface reconstruction.

\textbf{Co-located Light Assumption}. Decoupling a scene's material, geometry, and lighting conditions from the input multi-view images is not straightforward. To simplify the complexity of the task, we follow NeuralTO\cite{cai2024neuralto} to adopt a co-located lighting assumption, which assumes that there is only a point light located at the camera position and no ambient lighting. This lighting condition is similar to using a camera with a flash to take pictures of objects in a dark scene. 

\textbf{PBR with Deferred Rendering}. While some inverse rendering works based on 3DGS, such as RelightableGS\cite{relightablegs}, adopt the forward rendering method to implement the PBR module, we choose deferred rendering instead.

Deferred rendering is a classic real-time rendering technique, which brakes the properties of objects like normal and materials into texture maps and conducts complex lighting calculations in screen space. Deferred rendering has some advantages in our task. Firstly, pixel-level normal maps make the optimization process more stable \cite{3dgsdeferredreflection}, because Gaussian functions with correct normals transfer gradients to incorrect normals. 
Secondly, after obtaining the normal map and depth map, we can calculate the Fresnel reflectance and mix the specular, diffuse, and scattering components more conveniently. 
Meanwhile, deferred rendering does not require the calculation of a mixture for each Gaussian kernel, reducing the computational complexity.

We assign PBR related material properties to the surface Gaussians, including base color and roughness. Then, we obtain the material maps $\mathbf{M}_{\mathrm{attrib}}$ using 3DGS rendering pipeline:
\begin{equation}
    \mathbf{M}_{\mathrm{attrib}}= \sum_{i=1}^{N_{s}}{m_{\mathrm{attrib}}^i \alpha_i  \prod\limits_k^{i-1}{(1-\alpha_k)} }
\end{equation}
where $m_{\mathrm{attrib}}^i$ is the material property of each Gaussian, like base color or roughness. And then, wen calculate the specular reflection $C_{s}$ and diffuse reflection $C_{d}$ based on the diseny BRDF model\cite{disneybrdf},
\begin{align}
    \mathbf{C}_{\mathrm{s}} & = \frac{\mathbf{F}_{\mathrm{schlick}}\mathbf{D}_{\mathrm{GGX}}\mathbf{G}}{4(\mathbf{N} \cdot \mathbf{V})^2} \frac{\mathbf{L}_{\mathrm{cl}}}{\Phi(\mathbf{D})}, \\
    \mathbf{C}_{\mathrm{d}} & = \frac{\mathbf{M}_{\mathrm{baseColor}}}{\pi}\mathbf{F}_{\mathrm{schlick}}^2 \frac{\mathbf{L}_{\mathrm{cl}}} {\Phi(\mathbf{D})}
\end{align}
where, $\mathbf{N}$ is the normal map, $\mathbf{D}$ is the depth map, $\textbf{V}$ indicates the view direction for each pixel. $\mathbf{F}_{\mathrm{schlick}}$ denotes the Fresnel term with Schlick's approximation, $\mathbf{D}_{\mathrm{GGX}}$ denotes the GGX normal distribution function, $\mathbf{G}$ is the geometric function, and they are calculated from $\mathbf{N}$ and $\mathbf{V}$. $\mathbf{M}_{\mathrm{baseColor}}$ represents the albedo material map, $\mathbf{L}_{\mathrm{cl}}$ is the intensity of the co-located light source, and $\Phi(\mathbf{D})$ accounts for light attenuation based on distance. Under the co-located light assumption, the direction of light coincides with the view direction $\mathbf{V}$, allowing simplification of the denominator.

And then we render the scattering component $\mathbf{C}_{\mathrm{scatter}}$ using interior Gaussians. For one pixel $p$ in $\mathbf{C}_{\mathrm{scatter}}$, we denote the corresponding view direction as $w_o$, then the color for $p$ is:
\begin{equation}
    \mathbf{C}_{\mathrm{scatter}}(p) = \sum_i^{N_{i}} \mathrm{SH}_{\mathrm{scatter}}^i(w_o)\alpha_i  \prod\limits_k^{i-1}(1-\alpha_k)
\end{equation}
where $N_i$ is the number of the interior Gaussians the view ray passes through, and $\mathrm{SH}_{\mathrm{scatter}}^i$ represents the scattering color of a interior Gaussian kernel.

Finally, we incorporate the scattering component and account for its attenuation upon exiting the surface $(1 - \mathbf{F}_{\mathrm{schlick}})$. Thus, the final rendered result $\mathbf{C}_\mathrm{PBR}$ obtained by BSDF with deferred rendering can be expressed as:
\begin{equation}
    \mathbf{C}_{\mathrm{PBR}} = \mathbf{C}_{\mathrm{s}} + \gamma \mathbf{C}_{\mathrm{d}} + (1-\mathbf{F}_{\mathrm{schlick}})(\mathbf{C}_{\mathrm{scatter}})   
\end{equation}
where $\gamma$ is a hyperparameter used to adjust the transparency of surface, usually set to 0.3 in our experiments. Through PBR module, we can fully leverage depth maps and normal maps, thereby imposing more precise constraints on both surface normals and depth estimation.

\subsection{Training}
\textbf{Interior Gaussians Constraint}. Although supervision through rendering results can constrain surface Gaussians and interior Gaussians to some extent within the object silhouette, some internal Gaussians may still remain scattered outside the object surface. Under normal circumstances, interior Gaussians should be entirely enveloped by surface Gaussians. These stray interior Gaussians lead to incorrect ordering, which could potentially affect rendering quality or cause optimization instability. To ensure interior Gaussians remain properly contained within the object, we introduce a constraint term that drives externally located Gaussians toward the object surface:
\begin{equation}
    L_{\mathrm{in}} = \frac{1}{\mid G_{in} \mid}\sum_{i \in G_{in}} (1-sign(\Delta_{d_i})) \mid \Delta_{\mathbf{D}[p_x, p_y] - d_i} \mid 
\end{equation}
where $G_{in}$ denotes the set of interior Gaussians, $sign(x)$ equals 0 if $x \leq 0$ and equals 1 if $x > 0$. $(p_x, p_y)$ is the coordinate obtained by projecting the center of the interior Gaussian $G_i$ onto the screen space. $\mathbf{D}[p_x, p_y]$ denotes the value of the depth map $\mathbf{D}$ at pixel $(p_x, p_y)$, and $d_i$ is the depth of $G_i$ in view space. 

\textbf{Photometric Loss}. We employ L1 loss and SSIM loss to measure the discrepancy between the rendered images $C_{SH}$ and $C_{pbr}$ and the ground truth:
\begin{equation}
    L_{\mathrm{c}} = \lambda_1 (L_1^{\mathrm{SH}} + L_1^{\mathrm{PBR}}) + (1 - \lambda_1)(L_{\mathrm{SSIM}}^{\mathrm{SH}} + L_{\mathrm{SSIM}}^{\mathrm{PBR}})
\end{equation}

\textbf{Geometric Constraints for Surface Gaussians}. For geometric constraints, we adopt the single-view loss and multi-view geometric consistency loss proposed in PGSR. 
The complete geometric constraint can be expressed as:
\begin{equation}
    L_{\mathrm{surf}} = \frac{\lambda_2}{W} \sum_{p \in W} \parallel \mathbf{N}_d(p) - \mathbf{N}(p) \parallel_1 + \frac{\lambda_3}{V} \sum_{p_r \in V}  \parallel \mathbf{D}(p_r) - \mathbf{D}(p_n) \parallel_1
\end{equation}
where $W$ represents all pixels in the image, $\textbf{N}_d(p)$ denotes the normal computed from depth maps under local planar assumptions, $V$ indicates pixels with projection errors exceeding a threshold between consecutive frames, $p_r$ represents pixels in the current frame, and $p_n$ corresponds to matching pixels in adjacent frames.

\textbf{Constraints for PBR Materials}. To ensure proper decoupling of PBR material properties, we introduce smoothness regularization similar to RelightableGS to prevent abrupt variations in neighboring regions:
\begin{equation}
    L_{\mathrm{sm}} =  \parallel \nabla \mathbf{M}_{\mathrm{attrib}} \parallel e^{- \parallel \nabla C_{\mathrm{GT}} \parallel}
\end{equation}

Finally, the complete training objective combines all components:
\begin{equation}
    L = L_{\mathrm{c}} + L_{\mathrm{surf}} + \lambda_4 L_{\mathrm{in}} +  \lambda_5 L_{\mathrm{sm}}^{r} + \lambda_6 L_{\mathrm{sm}}^b
\end{equation}
where $L_{\mathrm{surf}}$ serves to flatten the surface Gaussians which is used in PGSR, $L_{\mathrm{sm}}^{r}$ and $L_{\mathrm{sm}}^{b}$ denote the smoothness constraints for the roughness and base color attributes, respectively.

\begin{figure}
    \centering
    \includegraphics[width=\linewidth]{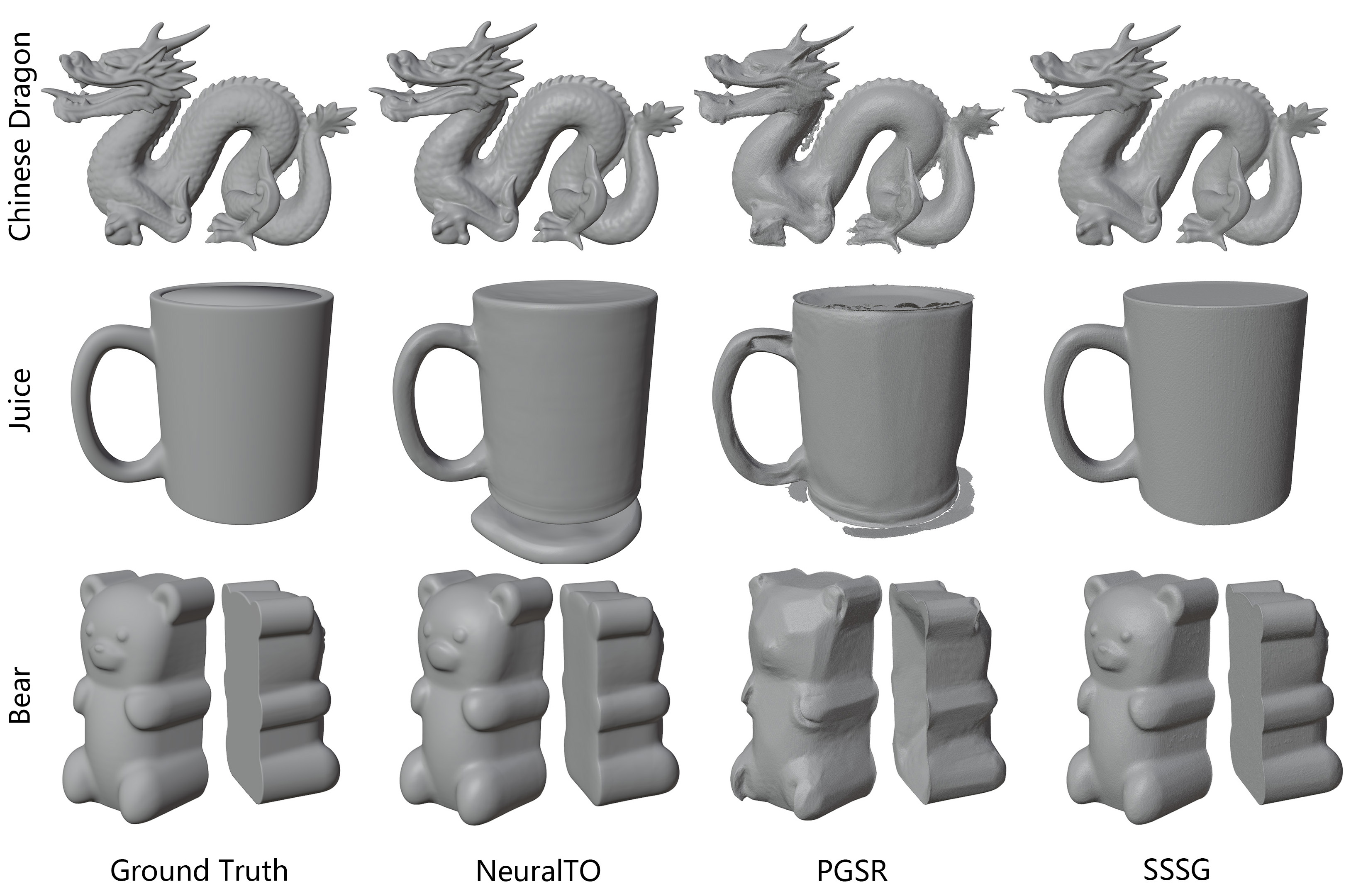}
    \caption{Surface reconstruction results on NeuralTO dataset. We compare several surface reconstruction method with our method, including NeuS-based (NeuralTO \cite{cai2024neuralto}) and 3DGS-based (PGSR\cite{chen2024pgsr}).}
    \label{fig:results}
\end{figure}
\section{Experiments}

\begin{table}[h]
\small
\caption{Quantitative results of chamfer distance(mm, $\times 10^{-3}$)$\downarrow$ on NeuralTO Syn dataset\cite{cai2024neuralto}. \label{tab:cd}}
\centering
\begin{tabular}{c | c c c c c c}
\hline
 & Yuanbao & C. Drag. & A. Drag. & Bear & Juice & Nail\\
\hline
NeuS\cite{neus} & \cellcolor{yellow!50} 8.5 & 64.6 & N/A & 4.7 & \cellcolor{yellow!50} 15.0 & 137.1\\
HFNeuS\cite{hf-neus} & 9.3 & 7.0 & 5.9 & 2.4 & 17.0 & 38.0\\
RefNeuS\cite{ref-neus} & 8.9 & \cellcolor{yellow!50} 4.5 & \cellcolor{yellow!50} 3.7 & \cellcolor{yellow!50} 2.3 & 21.6 & 32.7\\
NeuralTO\cite{cai2024neuralto} & \cellcolor{red!40} 1.2 & \cellcolor{red!40} 3.1 & \cellcolor{red!40} 1.0 & \cellcolor{orange!50} 1.1 & \cellcolor{orange!50} 13.2 & \cellcolor{orange!50} 2.2\\
2DGS\cite{2dgs} & 30.4 & 4.9 & 4.7 & \cellcolor{yellow!50} 2.3 & 25.3 & \cellcolor{yellow!50} 4.6\\
PGSR\cite{chen2024pgsr} & 19.4 & 6.7 & 17.4 & 9.9 & 23.4 & 39.5\\
Ours & \cellcolor{red!40} 1.2 & \cellcolor{orange!50} 3.7 & \cellcolor{orange!50} 2.2 & \cellcolor{red!40} 0.7 & \cellcolor{red!40} 3.3 & \cellcolor{red!40} 0.6\\
\hline
\end{tabular}
\end{table}

\textbf{Dataset}. We used the synthetic dataset provided by NeuralTO\cite{cai2024neuralto} to test the surface reconstruction and new perspective generation effects of our method. The dataset includes  six different translucent objects: ``Yuanbao'', ``Chinese Dragon'', ``Ancient Dragon'', ``Bear'', ``Nail'', and ``Juice''. This dataset uses the Principled BSDF shader in Blender to simulate the materials of translucent objects in a real environment. To test the adaptability of our algorithm, we also make some new scenes by rendering objects with different materials, which can be found in supplementary materials Section 2. For each scene, we uniformly sampled 120 viewing angles on the spherical surface. We also used the Principled BSDF shader for rendering. The rendering resolution was set to 800x800.

\textbf{Evaluation Criterion}. We used the Chamfer distance to evaluate the reconstruction quality. To evaluate the image quality of the new perspective generation, we used three widely used indicators, including PSNR (peak signal-to-noise), SSIM (structural similarity index measure), and LPIPS (the learned perceptual image patch similarity).

\textbf{Implementation Details}. COLMAP\cite{colmap} fails to reconstruct the initial points for 3DGS because of the lack of surface textures, so we adopted random point clouds to initialize the scene. Following the training strategy of PGSR, we only optimize the surface Gaussians in terms of photometric to achieve sufficient initialization in the first stage. In the second stage, we added the internal Gaussian and the geometric regularization terms. In our experiments, $\lambda_1$ is set to 0.2, $\lambda_3$, $\lambda_4$, and $\lambda_5$ are set to 0.01. $\lambda_{2}$ increases linearly from 0.0025 to 0.9. The training takes approximately 2.5 hours and consumes less than 8 GB of video memory. Same as \cite{newcombe2011kinectfusion}, we adopt the TSDF fusion algorithm as PGSR to extract the meshes. All experiments are conducted on an Nvidia RTX A6000.

\subsection{Surface Reconstruction Results}
We compare our method with the state-of-the-art surface reconstruction method NeuralTO\cite{cai2024neuralto}, and other methods including NeuS\cite{neus}, HF-NeuS\cite{hf-neus}, Ref-NeuS\cite{ref-neus}, 2DGS\cite{2dgs}, and PGSR\cite{chen2024pgsr}. We test these methods on the translucent objects dataset provided by NeuralTO. The quantitative results are shown in Table \ref{tab:cd}. Our method performs best on most scenes and has the lowest average Chamfer distance. 
From the visualization of results in Fig. \ref{fig:results}, we can find that our method performs better overall. Compared to normal 3DGS-based surface reconstruction methods like PGSR, our method has fewer obvious holes and more details. This demonstrates that our method improves the details of non-contour areas. More visualization of reconstructed results are provided in Supplementary Materials Section 1. 

\subsection{Novel View Synthesis Results}
\begin{table}
\small
\caption{Quantitative results of rendering quality for novel view synthesis on NeuralTO Syn dataset. \label{tab:nvs}}
\centering
\begin{tabular}{c | c c c c }
\hline
 & PSNR & SSIM & LPIPS\\
\hline
NeRF\cite{nerf} & 32.90 & 0.9605 & 0.0667\\
IRON\cite{iron} & 26.83 & 0.0947 & 0.7740\\
Ref-NeuS\cite{ref-neus} & 28.70 & 0.9562 & 0.0703\\
NeuralTO\cite{cai2024neuralto} & 41.34 & 0.9853 & 0.0232\\
3DGS\cite{3dgs} & \cellcolor{orange!50} 43.76 & \cellcolor{orange!50} 0.9946 & \cellcolor{orange!50} 0.0137\\
SSS\cite{sss} & 41.60 & 0.9917 & \cellcolor{yellow!50} 0.0143\\
GsShader\cite{gaussianshader} & 28.15 & 0.9327 & 0.0774\\
2DGS\cite{2dgs} & \cellcolor{yellow!50} 42.46 & \cellcolor{yellow!50} 0.9942 & 0.0167\\
PGSR\cite{chen2024pgsr} & 39.18 & 0.9917 & 0.0188\\
Ours & \cellcolor{red!40} 45.28 & \cellcolor{red!40} 0.9961 & \cellcolor{red!40} 0.0126\\
\hline
\end{tabular}
\end{table}

To verify the effect of our method on the new perspective generation task, we compared our method to NeRF\cite{nerf}, 3DGS\cite{3dgs}, SSS\cite{sss}, PGSR\cite{chen2024pgsr}, IRON\cite{iron}, GaussianShader\cite{gaussianshader}, and so on. We use $C_{\mathrm{SH}}$ as our rendering results in the evaluation. 
As shown in Table \ref{tab:nvs}, compared with NeuralTO and PGSR, our method has a significant improvement in rendering quality. This indicates that our method effectively resolves the optimization conflict between the appearance and the geometric regularization terms when dealing translucent objects. Compared with the NVS works, our method can still get similar or even better rendering effects. The visualization results are provided in Supplementary Materials Section 1.

\subsection{Ablation Study}
\begin{table}
\small
\caption{Ablation study on NeuralTO Syn dataset. \label{tab:ablation}}
\centering
\begin{tabular}{c | c c c c }
\hline
 & CD & PSNR & SSIM & LPIPS\\
\hline
w/o PBR & 0.0062 & 38.81 & 0.9929 & 0.0203\\
w/o Fresnel & 0.0075 & 39.91 & 0.9941 & 0.0164\\
w/o $G_{in}$ & 0.0086 & 40.90 & 0.9938 & 0.0179\\
full & 0.0020 & 45.28 & 0.9961 & 0.0126\\
\hline
\end{tabular}
\end{table}

\begin{figure}
    \centering
    \includegraphics[width=\linewidth]{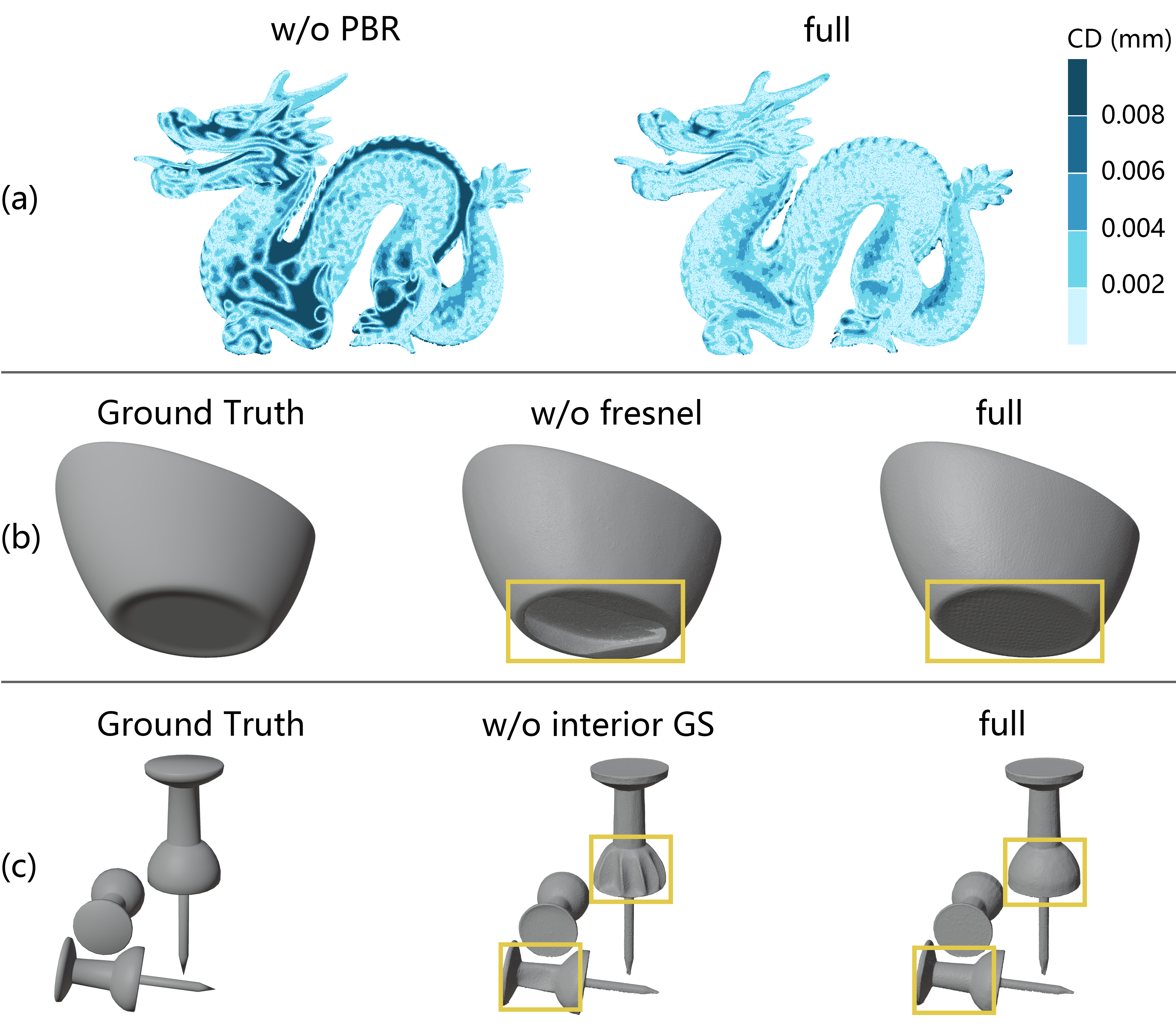}
    \caption{Comparison of different methods in the ablation study in some scenes of NeuralTO Syn dataset. (a) uses point wised error map to visualize the difference between wo PBR and full. (b) and (c) show the reconstructed meshes using different methods.}
    \label{fig:ablation}
\end{figure}

To verify the effectiveness of the new pipeline we proposed, we conduct ablation experiments from multiple aspects: (1) w/o PBR: estimate the geometry of the object surface without using PBR rendering; (2) w/o Fresnel: do not add the Fresnel term in the calculation of $C_{\mathrm{SH}}$; (3) w/o $G_{in}$: only use surface Gaussians. Specifically, since the distinction between surface and interior Gaussians is not made, the Fresnel term is no longer used for $C_{\mathrm{SH}}$, and the $\mathrm{SH}_{\mathrm{scatter}}$ property of the original interior Gaussians is changed to that of the surface Gaussians. (4) full: Add the Fresnel term and calculate $C_{\mathrm{PBR}}$ using deferred rendering. The comparison results are shown in Table \ref{tab:ablation} and Fig. \ref{fig:ablation}.

\textbf{PBR Rendering}. The optimization of PBR results, plays an important role in improving the accuracy of surface reconstruction. By decoupling lighting, surface geometry and object materials, it can improve the geometric structure accuracy of non-contour areas. It can be seen from Fig. \ref{fig:ablation} (a) that in the non-contour areas, the results after adding PBR module have better details. It can also be seen from Table \ref{tab:ablation} that the surface reconstruction quality has been significantly improved after the addition of the PBR module.

\textbf{Fresnel Term}. Table \ref{tab:ablation} proves that after adding the Fresnel term to explicitly represent the attenuation of light in and out of the object surface, the surface reconstruction results have been improved. It can be seen from Fig. \ref{fig:ablation} (b) that good rendering results can still be obtained without adding the Fresnel term, but it may lead to the optimization of the geometric structure in the wrong direction.

\textbf{Interior Gaussians}. The addition of interior Gaussians can help us to separate the surface color from the interior color well, thereby reducing the interference of interior scattering factors on the surface geometry during optimization and ensuring that the surface Gaussian can focus on representing the surface geometry structure. The results shown in Table \ref{tab:ablation} and Fig. \ref{fig:ablation} (c) prove the validity of our representation method, which combines surface Gaussians and internal Gaussians.

\subsection{Usage Samples for Real World Datas}
\begin{figure}
    \centering
    \includegraphics[width=\linewidth]{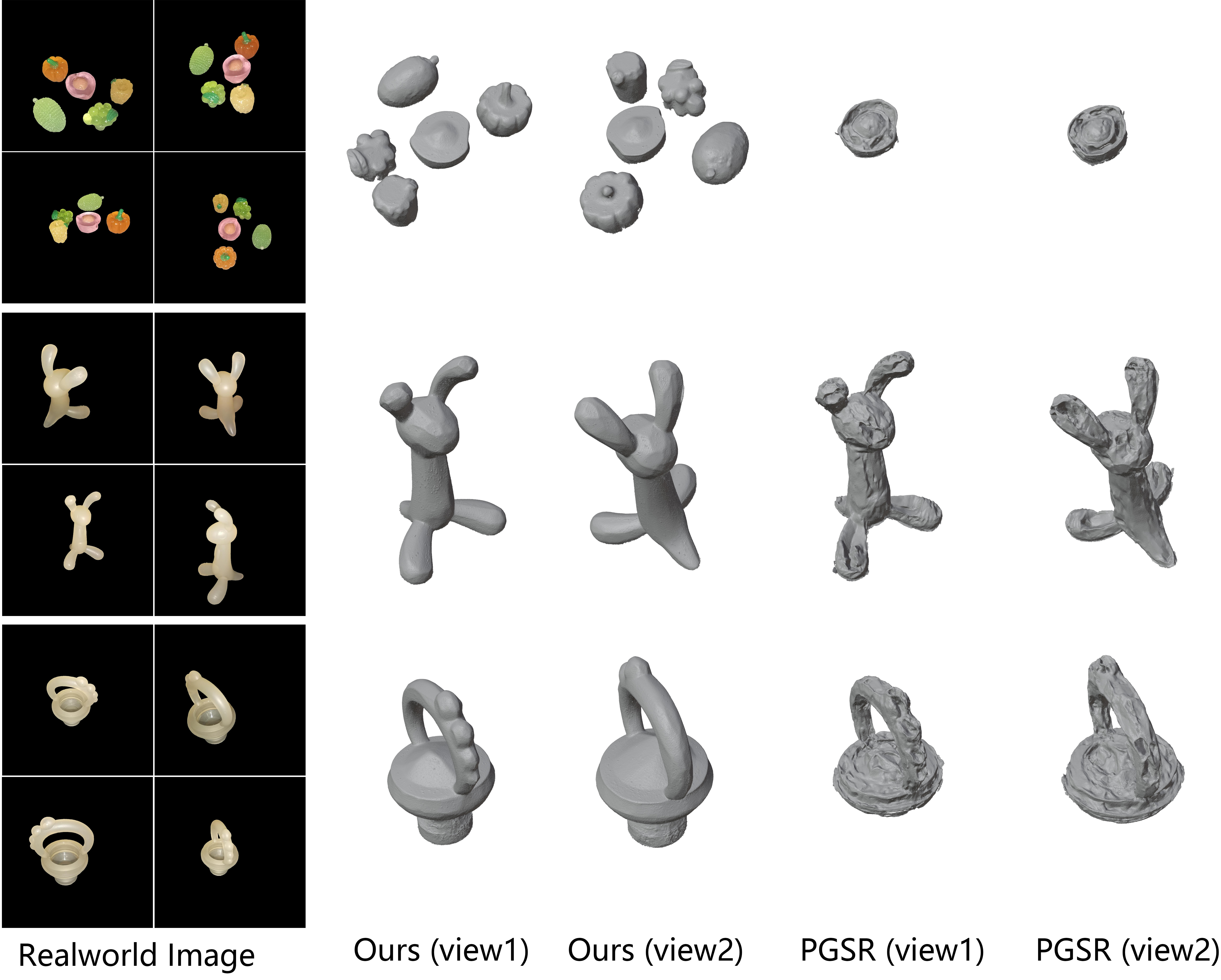}
    \caption{Surface reconstruction results on real-world translucent object datasets. The input images are captured by a camera with a flash in a dark environment.}
    \label{fig:real-world}
\end{figure}

We also tested our method on real-world translucent object datasets. The input images are captured by a camera with a flash in a dark environment. We test our method on 3 different real-world scenes. We capture about 30 images from different views. And we use SAM3 \cite{carion2025sam3segmentconcepts} to assist in segmentation and mask the background of the images with black. Since there is no ground truth for real-world datasets, we only provide some visualization results in Fig. \ref{fig:real-world}. It can be seen that our method can reconstruct relatively complete surface geometry.

\section{Discussion and Conclusion}

\textbf{Limitations.} Although PBR improves surface geometry, fine details (e.g., dragon scales) remain blurred due to insufficient material decoupling. Our model also simplifies light transport in translucent objects, omitting refraction and back-side illumination, which limits performance on transparent objects like glass. Future work may incorporate geometry priors, such as depth or normal estimation, to better handle transparency. Additionally, our method assumes co-located lighting, cannot handle environmental illumination, incorporating learnable environment maps may help.

We propose a novel pipeline for reconstructing translucent objects from multi-view images by combining surface and interior Gaussians to jointly model geometry and scattering. Surface Gaussians represent geometry and color, while interior Gaussians capture subsurface scattering. Incorporating the Fresnel term into 3DGS and introducing a Disney BSDF-based deferred rendering module improves surface reconstruction. Experiments on NeuralTO Syn show our approach is both more effective and efficient than existing methods.

\bibliographystyle{ACM-Reference-Format}
\bibliography{my-cite}

\end{document}